\documentclass[letterpaper,10pt,conference]{ieeeconf}    
\IEEEoverridecommandlockouts                             
\usepackage{graphicx}                                                    
\usepackage[hyphens]{url}
\usepackage{caption}
\usepackage{subcaption}
\usepackage{graphicx} % for pdf, bitmapped graphics files
\usepackage[hyphens]{url}
\usepackage{amsmath} % assumes amsmath package installed
\usepackage{multicol}
\usepackage{multirow}
\usepackage{rotating}

\title{\LARGE \bf
Near-optimal Keypoint Sampling for Fast Pathological Lung Segmentation}

\author{Awais Mansoor, Ulas Bagci$^*$, and Daniel J. Mollura% <-this % stops a space
\thanks{$^*$ indicates corresponding author. This work is supported by Center for Infectious Disease Imaging (CIDI), National Institutes of Health (NIH), Bethesda MD 20892.}% <-this % stops a space
\thanks{A. Mansoor, U. Bagci, and D. Mollura are with the Department of Radiology and Imaging Sciences, National Institutes of Health (NIH), Bethesda MD 20892.}
}

\begin{document}

\sloppy

\maketitle
\thispagestyle{empty}
\pagestyle{empty}

%%%%%%%%%%%%%%%%%%%%%%%%%%%%%%%%%%%%%%%%%%%%%%%%%%%%%%%%%%%%%%%%%%%%%%%%%%%%%%%%
\begin{abstract}

Accurate delineation of pathological lungs from computed tomography (CT) images remains mostly unsolved because available methods fail to provide a reliable generic solution due to high variability of abnormality appearance. Local descriptor-based classification methods have shown to work well in annotating pathologies; however, these methods are usually computationally intensive which restricts their wide-spread use in real-time or near-real-time clinical applications.  In this paper, we present a novel approach for fast, accurate, reliable segmentation of pathological lungs from CT scans by combining region-based segmentation method with local-descriptor classification that is performed on an optimized sampling grid. Our method works in two stages; during stage one, we adapted the fuzzy connectedness (FC) image segmentation algorithm to perform initial lung parenchyma extraction. In the second stage, texture-based local descriptors are utilized to segment abnormal imaging patterns using a near optimal keypoint analysis by employing centroid of supervoxel as grid points. The quantitative results show that our pathological lung segmentation method is fast, robust, and improves on current standards and has potential to enhance the performance of routine clinical tasks.
\end{abstract}

%%%%%%%%%%%%%%%%%%%%%%%%%%%%%%%%%%%%%%%%%%%%%%%%%%%%%%%%%%%%%%%%%%%%%%%%%%%%%%%%
\section{INTRODUCTION}
\begin{figure}[htb]
\centering
\includegraphics[scale =0.25]{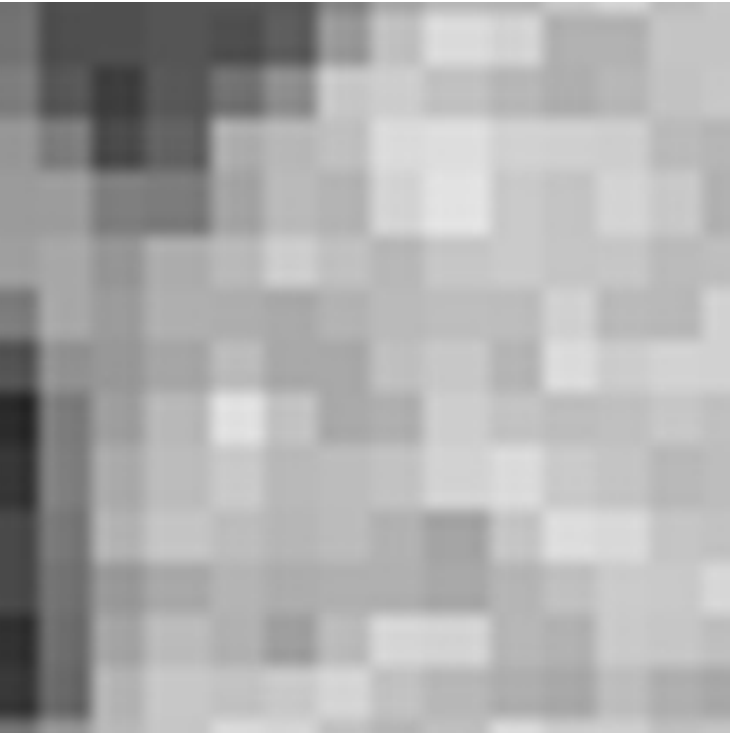}
\includegraphics[scale =0.25]{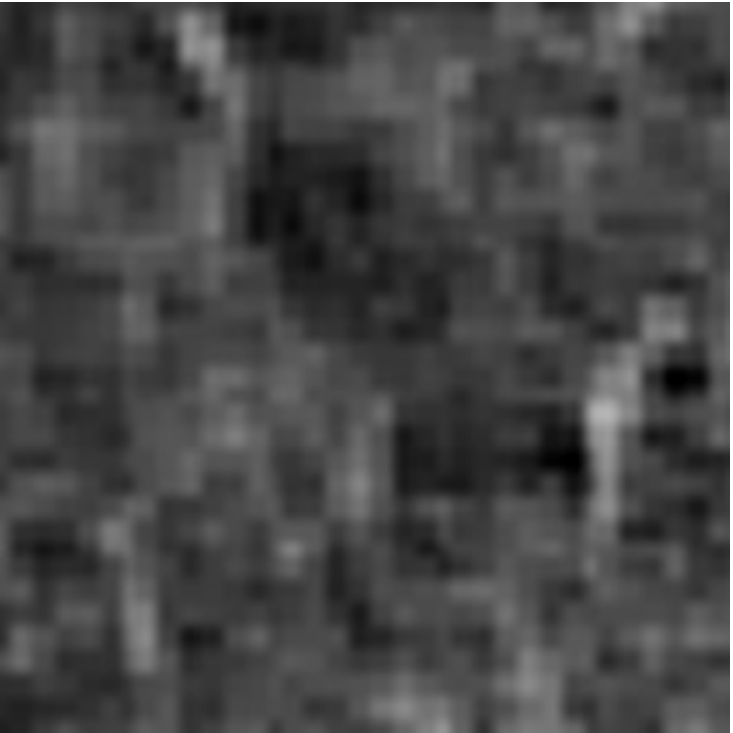}
\includegraphics[scale =0.25]{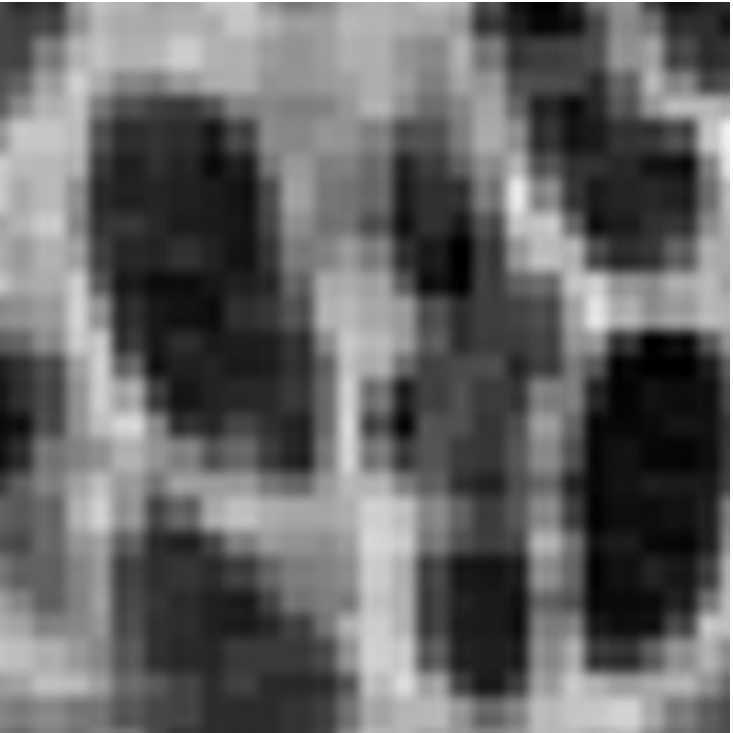}
\includegraphics[scale =0.25]{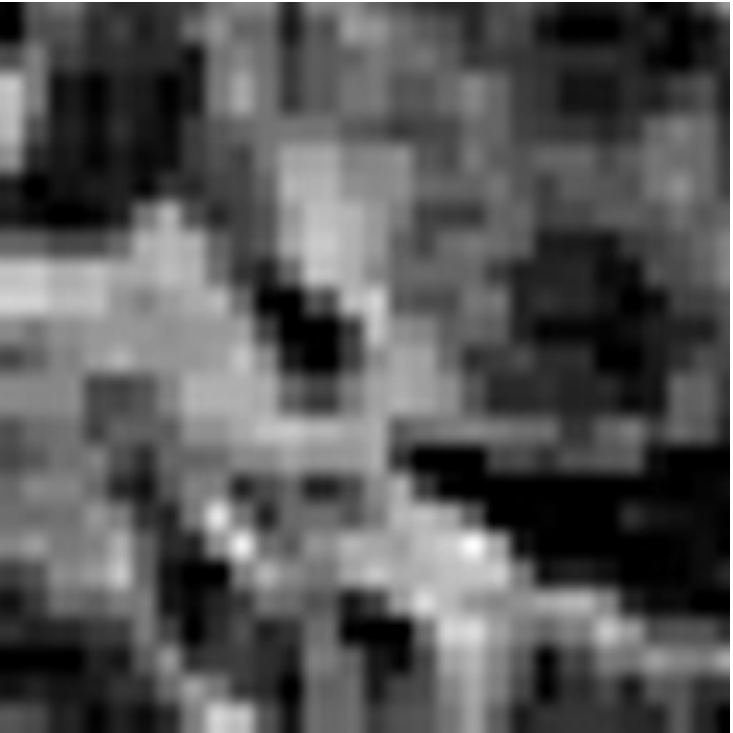}
\caption{\small{Schematic diagram showing most common abnormal texture patterns in pulmonary CT scans. \emph{from left to right}: consolidation, ground-glass-opacity (GGO), honey combing, and fibrosis.}}
\label{fig:texture}
\end{figure}

The use of different imaging modalities for assisting in diagnosis and quantification of disease has grown tremendously over the past decade \cite{xu2013559, mansoor2014statistical}. Specifically, CT images contain enormous amount of visual information (Fig. \ref{fig:texture}), making it impossible to analyze manually. In addition, imaging abnormalities often occur in small patches that can be better classified using local descriptors rather than global approaches because local descriptors are found to be extremely successful in detecting pulmonary abnormalities. However, the enormous amount of data to be processed in local descriptor calculations coupled with the trade-off between the accuracy and the efficiency, discourage their widespread use. Local descriptors can be applied in two modes in any image analysis task: dense-sampling and keypoint analysis. Dense sampling uses fixed grid to compute the local descriptors; however, the optimal density of the grid is not trivial to find: if the grid is too dense it will increase the computational cost by computing the local descriptors at redundant grid-points. On the other hand, if the grid is too sparse it will miss the clinically relevant information. \emph{Keypoint analysis} tries to find the optimal sampling grid for computing the local descriptors. The keypoint sampling has been widely used in various feature detection techniques in computer vision such as \emph{scale-invariant feature transform} (SIFT) and \emph{speed-up robust features} (SURF) to name a few.  

In this paper, we presented a novel approach to pathological lung segmentation using the keypoint sampling of supervoxels in CT scan. The near optimal grid for the keypoint sampling is created by adapting the state-of-the-art \emph{Simple Linear Iterative Clustering} (SLIC) method \cite{achanta20122274} to generate supervoxel for CT images. The modified SLIC has been demonstrated to generate texturally uniform supervoxels and adhering to the boundary constraints. The centroids of these texturally uniform atomic regions are used as keypoint sampling grid for the calculation of local descriptors. The performance evaluation tests conducted using surrogate truth obtained with the expert manual segmentation on a large dataset from diverse sources with varying amounts and types of abnormalities confirm the robustness, accuracy, and computational efficiency of our proposed method. 

\section{METHODS}
Our proposed algorithm consists of two steps. During the first step, initial segmentation of normal lung parenchyma is performed using region-based FC segmentation algorithm \cite{ciesielski2012375}. Second, the regions containing abnormal imaging patterns such as consolidation, GGO, etc. as well as the nearby soft tissues are subdivided into supervoxels and local descriptors at keypoint sampling points are calculated and classified using random-forest classifier to separate abnormal pulmonary regions from the neighboring soft tissues. The supervoxel classified as pulmonary pattern after classification are added to the initial FC segmentation to obtain annotated pulmonary volume. The details of the steps are presented below.

\subsection{Initial segmentation}
\begin{figure}[htb]
\centering
\includegraphics[scale =0.29]{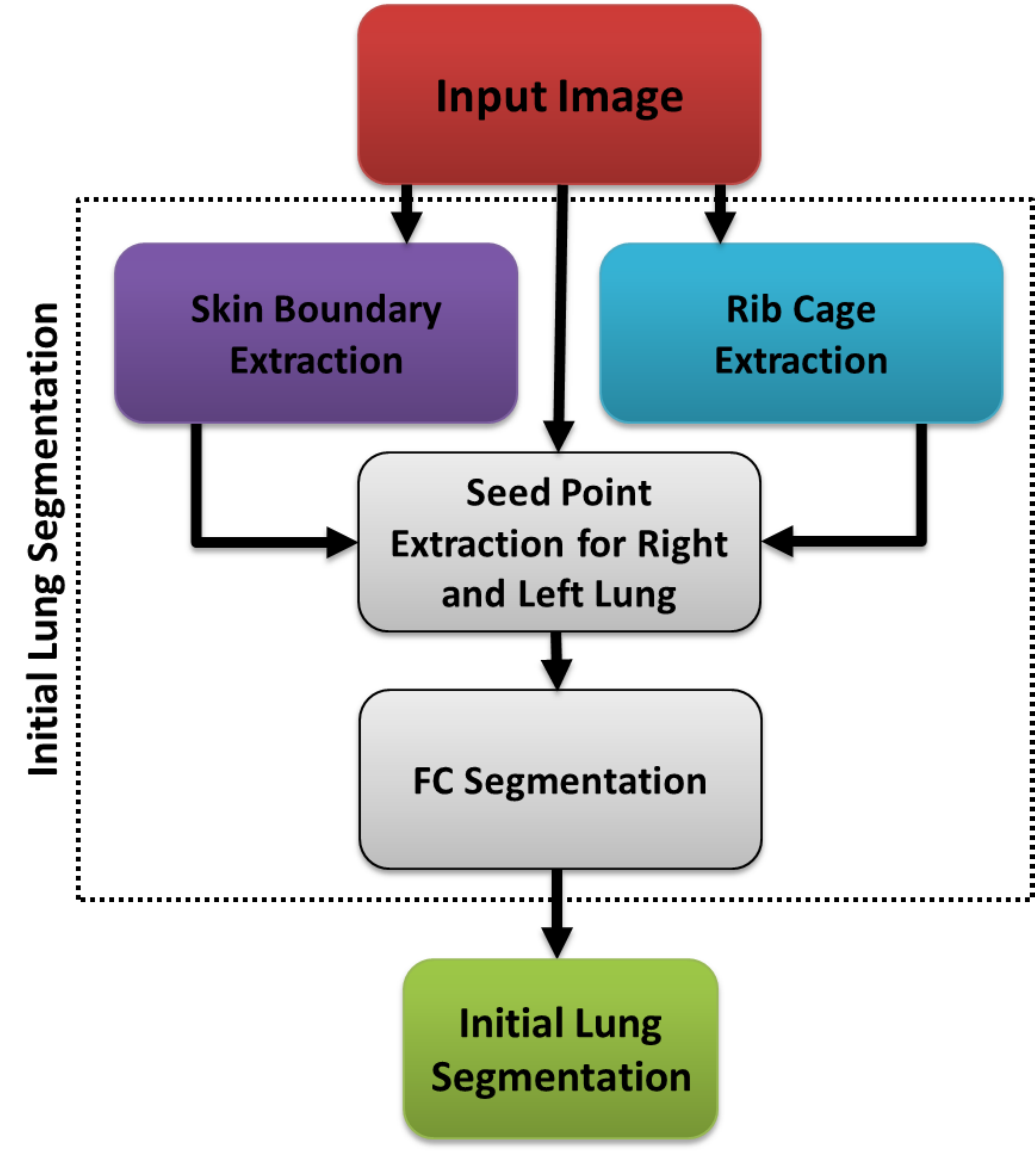}
\caption{\small{Flowchart explaining the initial segmentation using FC.}}
\label{fig:FC_Initial_Seg}
\end{figure}
Fig. \ref{fig:FC_Initial_Seg} summarizes the initial lung segmentation process using FC. FC requires two seed points: $s_l, s_r$, located within the left and right lungs, respectively.  In our design, we automatically set seed locations through a pre-processing step. To summarize, for any given CT image $I$, we use a thresholding operation $\mathcal{T}$ using CT attenuation values for normal lung parenchyma (Hounsfield Units (HU): -700 through -400, mean $\approx -550$ HU). Thus, $I^{\mathcal{T}}=\mathcal{T}\lbrace I \rbrace_{-550HU}$. Finally, we set the seed locations $s_l$ and $s_r$  after randomly sampling a few seed candidates (i.e., $3\times3\times3$ seed window) for each lung from $I^{\mathcal{T}}$ and select the voxels with minimum HU value as seeds:
\begin{eqnarray}
s_l \leftarrow  \mathcal{L} (\min_{HU} ROI^{3\times3\times3}_{random}) \in I^{\mathcal{T}}_{left}, \nonumber \\
s_r \leftarrow  \mathcal{L} (\min_{HU} ROI^{3\times3\times3}_{random}) \in I^{\mathcal{T}}_{right}.
\end{eqnarray}
where $\mathcal{L} $ denotes the location of the voxel(s), and $I^{\mathcal{T}} = I^{\mathcal{T}}_{left} \cup I^{\mathcal{T}}_{right}$.

Apart from seeds, FC algorithm requires approximate mean $m$ and the standard deviation $\sigma$ of the lung region to be used in affinity functions. These values were empirically set to normal lung parenchyma as $m=-550$ HU, $\sigma=150$ HU after analyzing hundreds of CT images. Once seeds and affinity parameters for FC are set, the initial delineation is performed. The output of the FC segmentation is a binary mask of the lung fields. The extent of how well the initial FC segmentation performs depends on the amount, kind, and density of abnormality present in the target image. 

\subsection{Abnormality detection}
\begin{figure}[htb]
\centering
\includegraphics[scale =0.23]{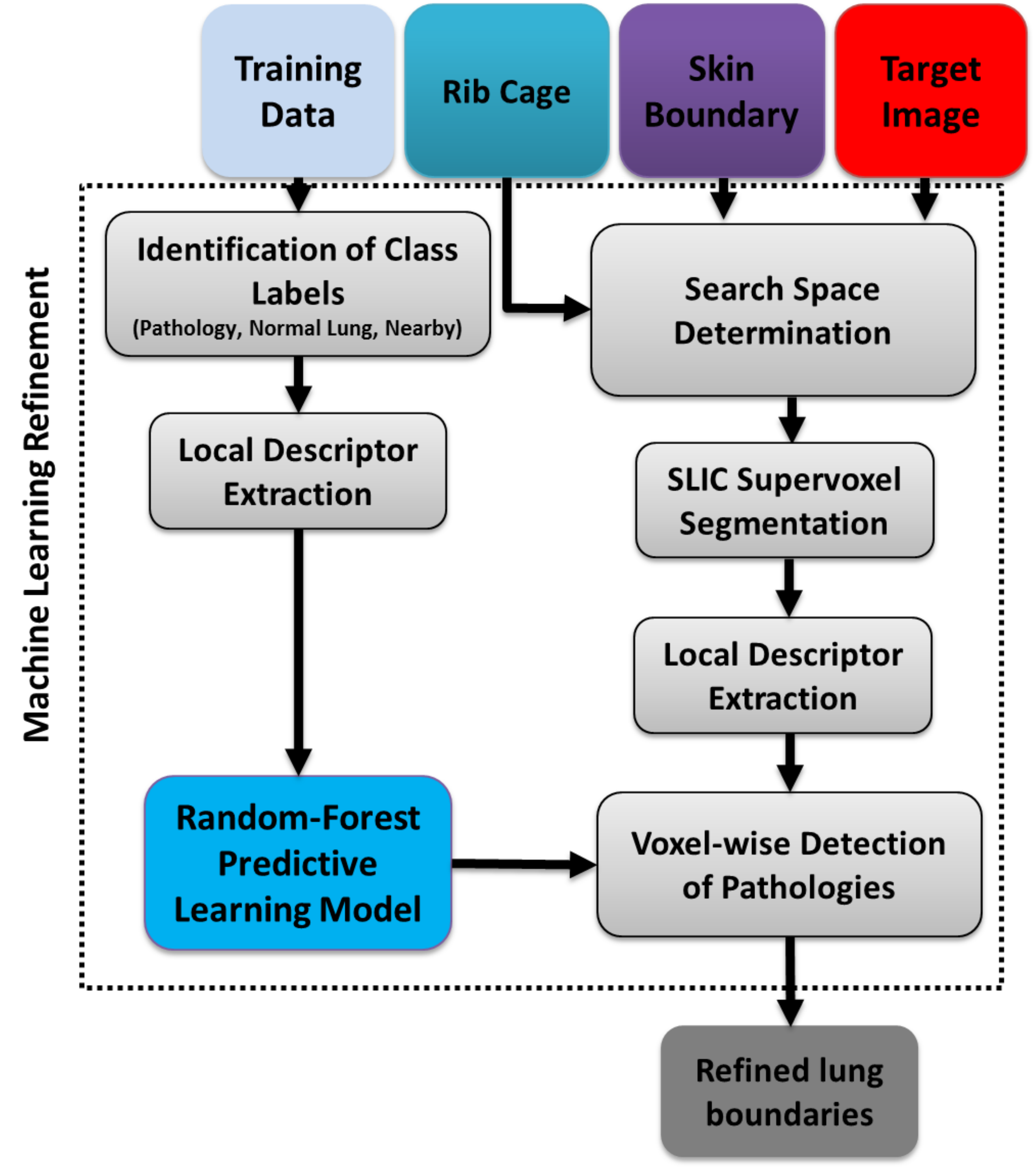}
\caption{\small{Schematic diagram showing texture classification-based refinement, highlighting both \emph{predictive model learning} for RF and the \emph{pathology classification} in the target image based on the learned predictive model.}}
\label{fig:machine_learning}
\end{figure}

Abnormality detection module consists of two parts. First, a search-space for prospective abnormalities is defined that includes the nearby soft tissue as well as abnormal pulmonary areas that have intensities very similar to nearby soft tissue. The search space is subsequently partitioned into supervoxels by applying an adapted SLIC algorithm to handle gray-scale 3D volumetric CT scans. The algorithm partitions the CT image into superpixels of nearly uniform sizes whose boundaries closely match with the natural boundaries in the image while capturing the redundancy in the data \cite{fulkerson2009670}. Supervoxels provide a primitive neighborhood from which a single representative local descriptor can be computed for all voxels belonging to that supervoxel. Finally, the random forest classifier is used for binary classification (pulmonary abnormality, near-by soft tissue) of the descriptors. Fig. \ref{fig:machine_learning} summarizes the abnormality segmentation module, the details are presented in the following subsections.

\subsubsection{SLIC supervoxel segmentation}
In \cite{achanta20122274} Achanta et al. introduced a memory efficient method named, \emph{simple linear iterative clustering}, for generating superpixels and supervoxels. The method has since then been applied in various applications and has shown to exhibit excellent preservation; therefore, improving the overall performance of the subsequent delineation method. The only parameter to the algorithm is the desired number of approximately equal-sized supervoxels $k$. The clustering process begins inside a 4-dimensional space where $k$ cluster centers $C_i=\{v, x, y, z\}_i,\forall i\in\{1,\dots,k\}$ are sampled on a regular grid. To produce roughly equal-sized supervoxel the grid interval $S$ is set to $S = \sqrt {\frac{N}{k}}$, where $N$ is the total number of supervoxels. Next, each voxel is assigned to the nearest cluster center whose search space coincides with the voxel location. Once every voxel has been associated with a cluster center, an update step is performed to adjust the cluster center to the mean of $[v,x,y,z]_i$ vector of all voxels belonging to the cluster $i$. The $L_2$ norm is then used to estimate the residual error between the previous and the updated cluster center locations. The update step is repeated iteratively until the error converges.

\subsection{Local descriptor extraction} 
\begin{figure}[htb]
\centering
\includegraphics[scale =0.25]{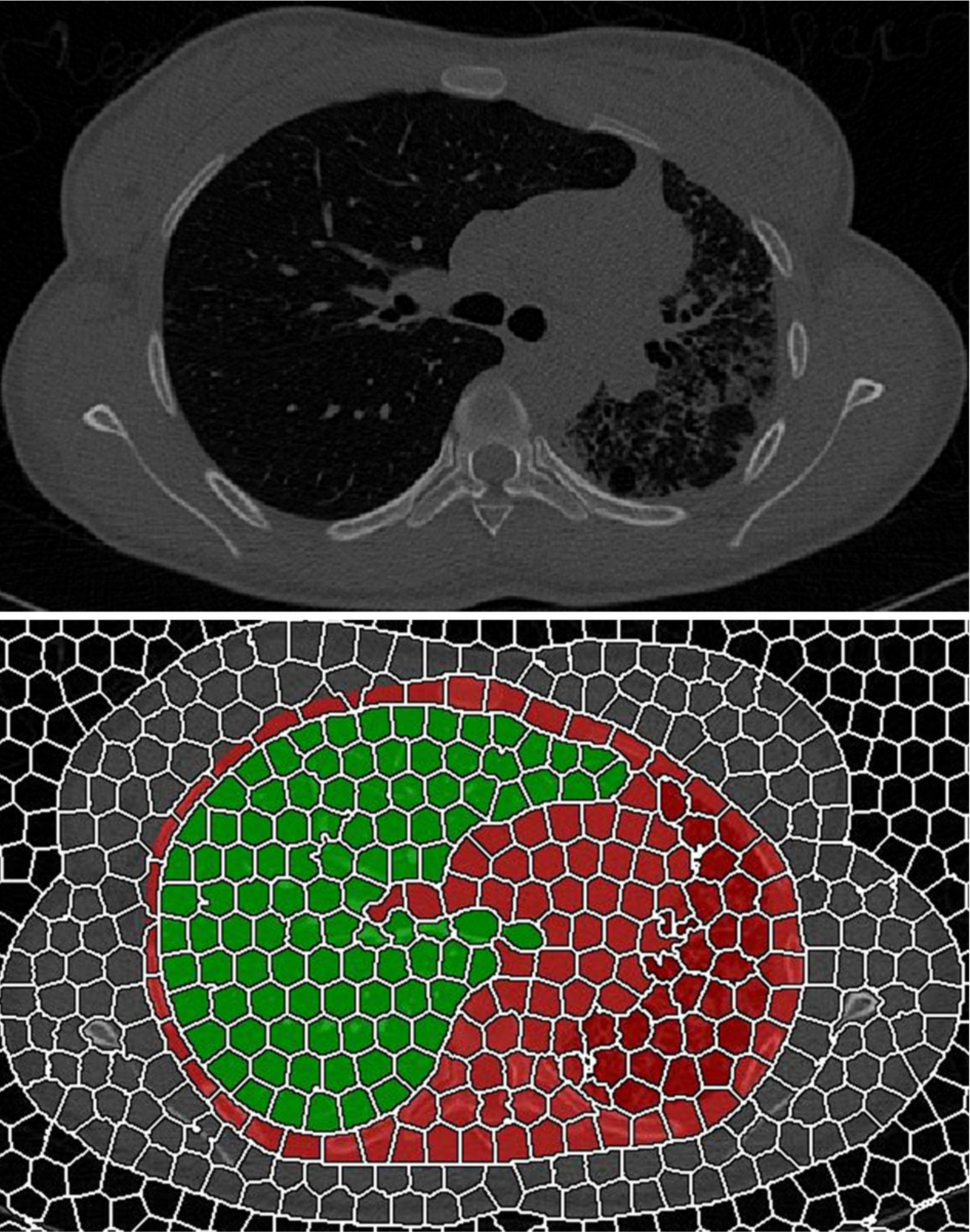}
\caption{\small{Schematic diagram showing the supervoxel-segmented search space (\emph{red}) with initial FC segmentation (\emph{green}).}}
\label{fig:searchSpace}
\end{figure}

\begin{figure}[htb]
\centering
\includegraphics[scale =0.24]{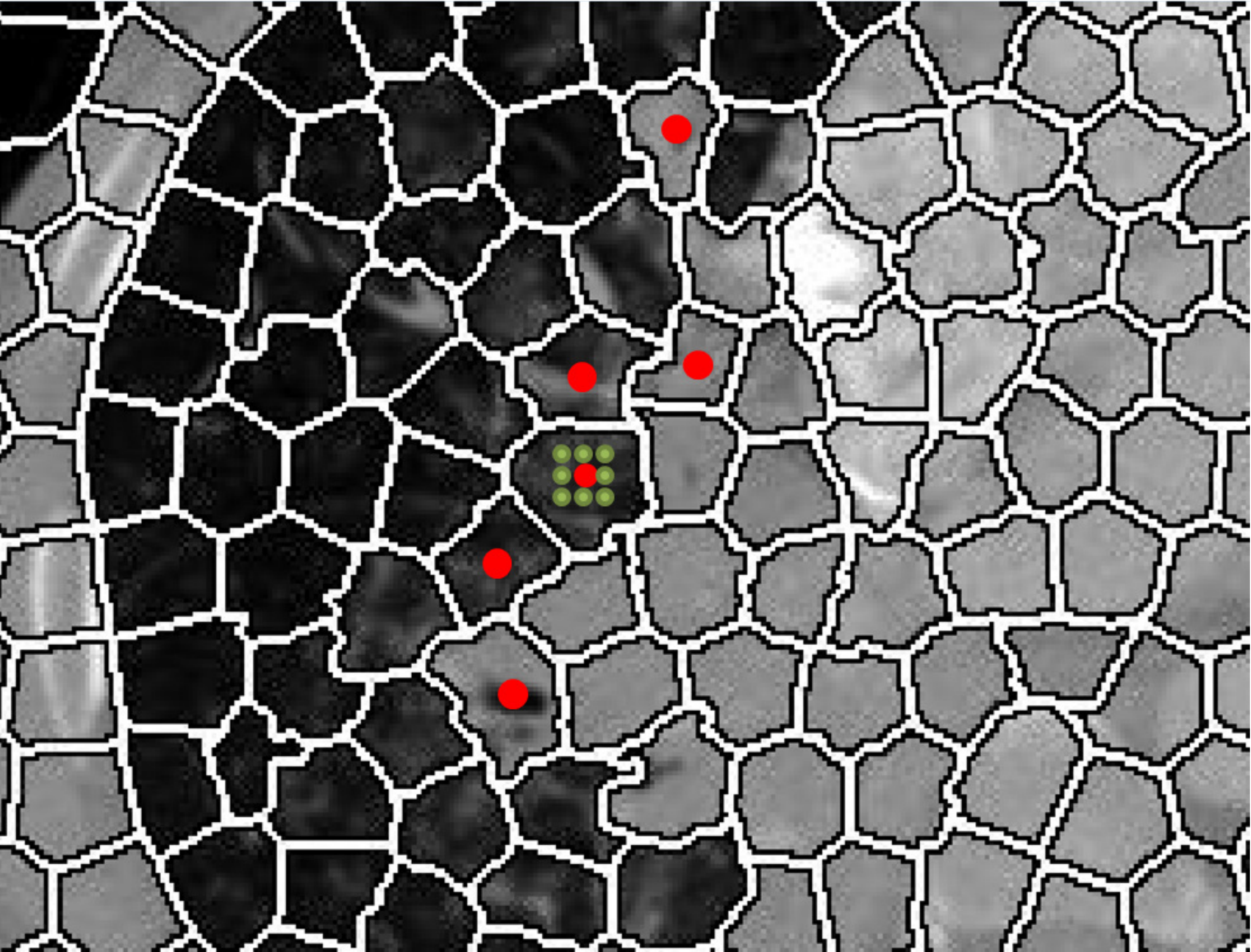}
\caption{\small{Schematic diagram showing the centroid of a supervoxel (\emph{red dots}) with neighborhood (\emph{green dots}) for local descriptor extraction.}}
\label{fig:centroid}
\end{figure}

Although local descriptors are very useful for detecting local patterns such as pathologies, it is not trivial to extract discriminative feature sets to drive the detection process. Moreover, assessing every voxel's class label may be computationally expensive. To address these two challenges, we integrated rib cage extraction and convex-hull fitting processes into the random forest classification algorithm in order to restrict the search space (Fig. \ref{fig:searchSpace}) to rib cage area only. To further reduce the redundancy, the local descriptors are calculated only at the centroid of the supervoxels (Fig. \ref{fig:centroid}) within the search space forming an optimized keypoint sampling grid since the centroid of texturally uniform supervoxel can be assumed to representative of the entire supervoxel. For random forest voxel classification of lung tissues, we employ \emph{grey-level run length matrix} (GLRLM), \emph{gray-level co-occurrence matrix} (GLCM), and histogram features. Justification of the use of GLCM, GLRLM, and histogram features is based on the visual analysis of CT lung pathologies~\cite{Song2013797}. 

For every voxel within the search region, we extract the features considering a region-of-interest around the voxel. The complete list of $24$ distinct features is shown in Table~\ref{table:classification_features}.
\begin{table}[!t]
%% increase table row spacing, adjust to taste
\renewcommand{\arraystretch}{1.2}
\caption{\small{Extracted features for voxel-wise classification of lung tissues.}}
\label{table:classification_features}
\centering
\begin{tabular}{c|l}
		  \hline
      \multirow{8}{*}{\begin{sideways}\textbf{GLCM}\end{sideways}} &Energy\\ & Entropy\\ &Correlation\\ &Inverse Difference Moment (IDM)\\ & Inertia \\ & Cluster Shade (CS) \\& Cluster Prominence (CP) \\
			\hline
			\multirow{13}{*}{\begin{sideways}\textbf{GLRL}\end{sideways}} &Short Run Emphasis (SRE) \\ & Long Run Emphasis (LRE) \\ &  Gray-Level Non-uniformity (GLN) \\ & Run Length Non-uniformity (RLN)\\ & Run Percentage (RP) \\& Low Gray-Level Run Emphasis (LGRE) \\& High Gray-Level Run Emphasis (HGRE)\\ & Short Run Low Gray-Level Emphasis (SRLGE)\\ & Short Run High Gray-Level Emphasis (SRHGE) \\ & Long Run Low Gray-Level Emphasis (LRLGE)\\ & Long Run High Gray-Level Emphasis (LRHGE)\\ 
			\hline
			\multirow{2}{*}{\begin{sideways}\textbf{ Hist.}\end{sideways}}&\begin{tabular}{llll} Mean & Variance & Skewness\\
			 Kurtosis & Min.  & Max. \\
			\end{tabular}
			\\
			\hline
    \end{tabular}
\end{table}

Since, our aim in this study is only to segment pathological lungs, we considered all pathological regions not captured by FC as belonging to a single label (i.e., $T_p$). All the voxels in the search space, $\mathcal{R}_{ss}$, are classified into two classes: pathological ($T_p$) or non-pathological ($T_n$) regions. In particular, neighboring structures of the lung were considered as non-lung and/or non-pathological structures and labeled as $T_n$. 

Random forest has been shown to be powerful for  various pattern classification tasks due to its high accuracy, efficiency, and robustness. In training a random forest classifier, two experienced observers annotated various pathology patterns from randomly selected CT scans (21 CT scans from different subjects). A total of 997 non-overlapping ROIs were extracted from those annotations such that 507 observations belong to $T_p$ while 490 observations belong to $T_n$. A random forest classification model is constructed using those observations with the corresponding labels. Table~\ref{table:classification_parameters} summarizes the set of parameters used for feature extraction and random forest classifier training.
\begin{table}[!t]
\renewcommand{\arraystretch}{1.3}
\caption{\small{Parameter settings for random forest classification method in pathology detection.}}
\label{table:classification_parameters}
\centering
\begin{minipage}[t]{0.48\linewidth}
\begin{tabular}{c|p{3.1cm}}
      \hline
      \multirow{4}{*}{\begin{sideways}\textbf{GLCM}\end{sideways}} & $\#$ of bins per axis = 16 \\ & $\#$ of directions = 4\\ & Offset = 2 \\ & Pixel intensity dynamic range = 16 bits\\			
			\hline
			\multirow{2}{*}{\begin{sideways}\multirow{2}{*}{\textbf{Rnd. fst.}}\end{sideways}} & $\#$ of trees in a forest = 70\\ & $\%$ of training set used to build
                    individual trees = 0.6\\
			\hline			
    \end{tabular}
\end{minipage}
\begin{minipage}[t]{0.48\linewidth}
\begin{tabular}{c|p{3.1cm}}
      \hline      
			\multirow{4}{*}{\begin{sideways}\textbf{GLRLM}\end{sideways}}&\\ & $\#$ of directions = 4\\ & $\#$ of levels = 8\\&\\			
			\hline
			\multirow{3}{*}{\begin{sideways}\textbf{Misc.}\end{sideways}}  &ROI dynamic-range = 16-bits\\ & ROI window = $7\times 7$\\&\\
			\hline
    \end{tabular}
\end{minipage}
\end{table}

\section{Performance Evaluation}
\begin{table}[htb]
\renewcommand{\arraystretch}{1.3}
\caption{Data Description of the CT scans used in our experiments.\label{table:data_descriptors}}
\centering
\begin{tabular}{p{1cm} c c p{3cm}}
\hline
& \textbf{\# of patients} &\textbf{\# of scans} &\textbf{Notes}\\
\hline
\textbf{Tuberculosis} & 80 & 93 &  Patient scans with TB-infected lungs. The data contains various pathologies prominently: consolidation, GGO, cavity, bleb, and pleural effusion. (\url{http://tuberculosis.by/}).\\
\textbf{LOLA} & 55& 55&Data set designed as part of the lung segmentation challenge (MICAII 2011). The pathologies in lung scans range from mild to severe. (\url{http://lola11.com}).\\
\textbf{EXACT} &40 & 40& High-resolution scans with mild pathologies (consolidation, and effusion). (\url{http://image.diku.dk/exact}).\\
\hline
\textbf{Total}&175& 188 & \textemdash\\
\hline
\end{tabular}
\end{table}
To evaluate the performance of our proposed method, we used publicly available datasets from various sources. The description of the evaluation data is provided in Table \ref{table:data_descriptors}. When biopsy images are not available, manual expert evaluation is mostly accepted as the gold standard. For reference standard in our study, the reference standard were provided by two experienced observers through manual segmentation. Dice similarity coefficient (DSC) was used as evaluation metrics for quantitative analysis. Table \ref{table:results} summarizes the quantitative evaluation of the annotation produced by our method. Results clearly demonstrates the effectiveness of the proposed technique in dealing with most commonly encountered pathologies.
\begin{table}[htb]
\renewcommand{\arraystretch}{1.3}
\caption{\small{Overall performance of the proposed lung segmentation approach averaged over different data sets and averaged overall. Mean and standard deviation (std) is provided for each index.}}
\label{table:results}
\centering
\begin{tabular}{l p{1cm} p{1.4cm} p{1.5cm}}
\hline
 \textbf{Data set}& &\multicolumn{2}{c}{\textbf{DSC}}\\
\hline
& &\textbf{Observer-I} &\textbf{Observer-II} \\
\hline
\textbf{Tuberculosis Data}& mean std & 0.9611 0.0326&0.9692 0.0133\\
\textbf{Exact}& mean std& 0.9801 0.0121&0.9797 0.0143\\
\textbf{Lola}& mean std&0.9527 0.0149& 0.9623 0.0223\\
\hline
\textbf{Total}& mean std&0.9527 0.0149& 0.9623 0.0223\\
\hline
\end{tabular}
\end{table}
\section{Conclusion}
In this paper, we present a fully automated lung segmentation method for CT scans with and without pulmonary abnormalities. The distinctive pulmonary abnormality patterns are mostly localized; therefore, global region-based methods mostly fail to capture those abnormalities. Local descriptors are found to extremely successful in finding those abnormal patterns; however, these methods are computationally expensive, therefore, efficient computation of local descriptors requires appropriate grid density estimation which is difficult to estimate. To address this issue, we propose a novel keypoint sampling method based on texturally uniform atomic regions known as \emph{supervoxels}. Our method begins with computationally efficient initial FC segmentation to segment the normal lung regions, followed by a machine-learning-based local refinement scheme performed on near-optimal key-point sampling grid formulation to capture missing abnormal patterned pulmonary regions. The evaluation results shows an overlap score (DSC) of greater than 95\% from publicly available challenge data sets, indicating a successful pathological lung segmentation system to be used in routine clinics. 
\bibliographystyle{IEEEtran}
\bibliography{IEEELung} % biography section
\end{document}